\documentclass[runningheads]{llncs}

\usepackage{eccv}

\usepackage{eccvabbrv}

\usepackage{graphicx}
\usepackage{float}
\usepackage{booktabs}

\usepackage[accsupp]{axessibility}  

\usepackage{array}
\usepackage{fontawesome5}
\usepackage{float}
\usepackage{multirow}
\usepackage{arydshln} 
\usepackage[dvipsnames]{xcolor}

\usepackage{pgf}
\usepackage{pgfplots}

\everymath=\expandafter{\the\everymath\displaystyle}
\makeatletter\@ifpackageloaded{underscore}{}{\usepackage[strings]{underscore}}

\makeatother
\usepackage{tikz}
\usepackage{pgfplots}
\pgfplotsset{compat=1.14}

\usetikzlibrary{shapes}
\usetikzlibrary{calc}
\usetikzlibrary{backgrounds}
\usetikzlibrary{positioning}
\usetikzlibrary{arrows}
\usetikzlibrary{arrows.meta}
\usetikzlibrary{decorations.text}    
\usetikzlibrary{fit}
\usetikzlibrary{spy}
\usetikzlibrary{3d}
\usetikzlibrary{positioning}
\usepackage[weather]{ifsym}

\usepackage[skins]{tcolorbox}

\definecolor{turquoise}{cmyk}{0.65,0,0.1,0.3}
\definecolor{purple}{rgb}{0.65,0,0.65}
\definecolor{dark_green}{rgb}{0, 0.5, 0}
\definecolor{light_green}{rgb}{0.2, 0.7, 0.2}
\definecolor{orange}{rgb}{0.7, 0.5, 0.5}
\definecolor{orangeish}{rgb}{0.8, 0.7, 0.1}
\definecolor{ipadaptercolor}{rgb}{0.8, 0.7, 0.1}
\definecolor{red}{rgb}{0.8, 0.2, 0.2}
\definecolor{darkred}{rgb}{0.6, 0.1, 0.05}
\definecolor{blueish}{rgb}{0.0, 0.3, .6}
\definecolor{light_gray}{rgb}{0.7, 0.7, .7}
\definecolor{dark_gray}{rgb}{0.3, 0.3, .3}
\definecolor{pink}{rgb}{1, 0, 1}
\definecolor{greyblue}{rgb}{0.25, 0.25, 1}

\usepackage{hyperref}

\usepackage{orcidlink}

\title{Stylecodes: Encoding Stylistic Information For Image Generation}
\titlerunning{Stylecodes}

\author{Ciara Rowles}

\authorrunning{C.~Rowles}

{\tt\small }
\institute{
Corresponding author: \href{mailto:crowles98@gmail.com}{crowles98@gmail.com}
}

\usepackage[width=122mm,left=12mm,paperwidth=146mm,height=193mm,top=12mm,paperheight=217mm]{geometry}
\begin{document}
\maketitle
\begin{figure}
    \centering
    \includegraphics[width=0.65\linewidth]{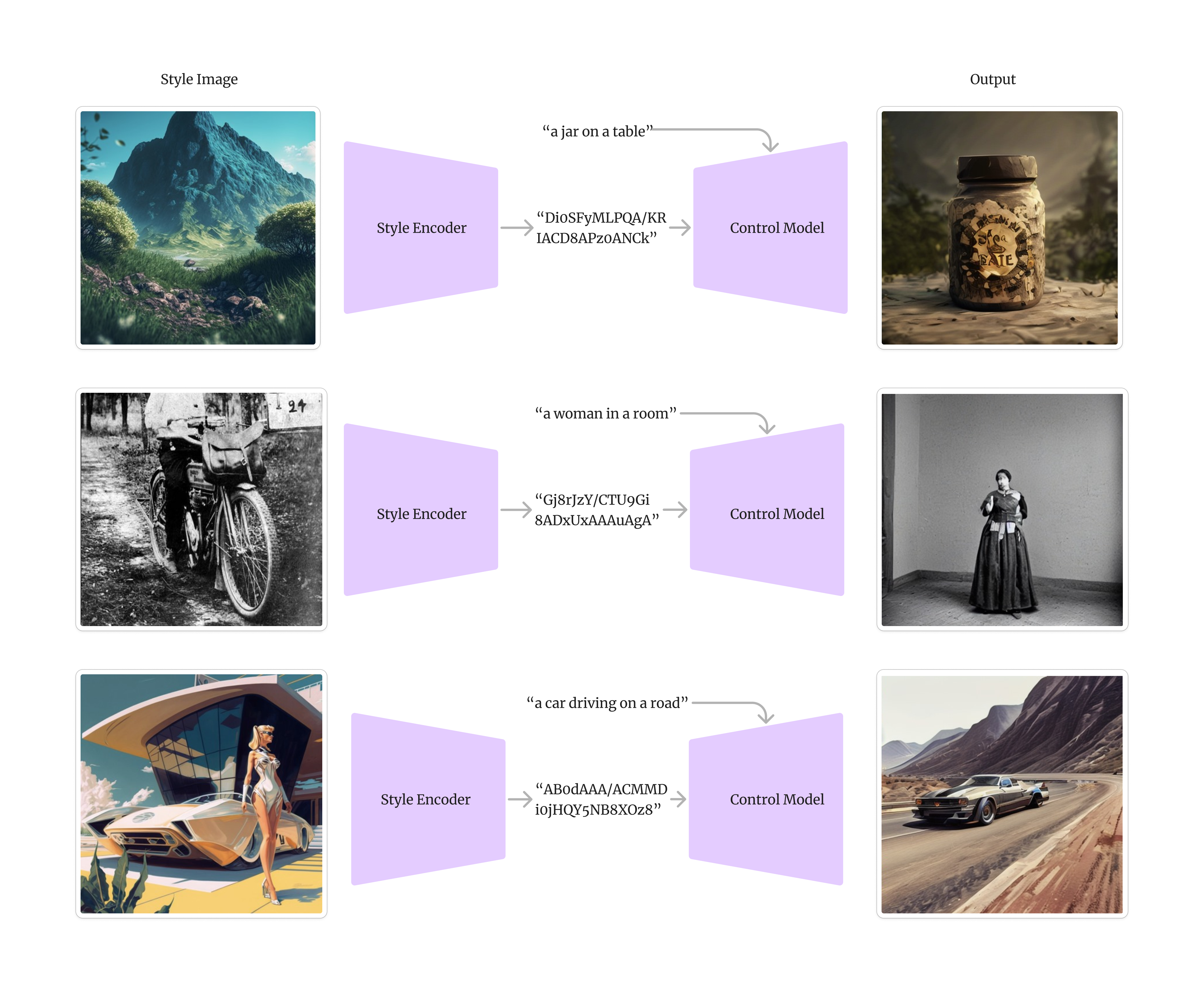}
    \caption{Our Style Encoder compresses image styles into compact strings for style-conditioned generation.}
\end{figure}
\begin{abstract}
Diffusion models excel in image generation, but controlling them remains a challenge.
We focus on the problem of style-conditioned image generation.
Although example images work, they are cumbersome: srefs (style-reference codes) from MidJourney solve this issue by expressing a specific image style in a short numeric code.
These have seen widespread adoption throughout social media due to both their ease of sharing and the fact they allow using an image for style control, without having to post the source images themselves.
However, users are not able to generate srefs from their own images, nor is the underlying training procedure public.
We propose StyleCodes: an open-source and open-research style encoder architecture and training procedure to express image style as a 20-symbol base64 code.
Our experiments show that our encoding results in minimal loss in quality compared to traditional image-to-style techniques.
\end{abstract}
\section{Introduction}\label{sec:intro}%

The field of image generation has received fresh impetus from diffusion model theory, where it is seen as an iterative process that reverses the diffusion of images into pure noise.
Diffusion models have achieved state-of-the-art performance in image generation tasks and have proven to be more robust to train than the previous GAN-based models.
They are also more widely applicable in practice, in large part because of the classifier-free guidance technique, which offers unprecedented control over the output image using a text prompt~\cite{CFG}.

However, the common proverb ``An image is worth a thousand words'' is as relevant as ever: crafting text prompts that result in exactly the desired image is a non-trivial task, referred to as \emph{prompt engineering}.
Even when the desired image is clear in the mind of the user, it is difficult to express it in such a way that the model output replicates it accurately; expressing intent through images is often far more intuitive than through text.
This concept has spurred the development of image-based conditioning such as ControlNet\cite{ControlNet} and IPAdapter\cite{IPAdapter, IPAdapterInstruct} approaches.
From scribbles and sketches or stylistic examples to many other modalities: these methods allow users to express their intent in the image domain, where spatial information and stylistic cues can be more easily conveyed.

Yet in practice, these control methods lack controllability and the ability to collaborate broadly without extensive shared workflows.
We tackle this problem by first creating "stylecodes", 20 digit base64 encoded codes with a combination of a custom encoder and a custom Stylecode-conditioned model  for control of the UNet, in this case with Stable Diffusion 1.5~\cite{SD15}.
We outline how to both encode and decode a stylecode and how this can be used to generate stylized images in normal contexts, as well as how the complete architecture is trained.

\section{Related Work}\label{sec:related}

\subsection{Diffusion Models}

Diffusion models generate images conditioned on text by learning to reverse a gradual diffusion process~\cite{DiffusionTheory,DDPM,DDIM}, typically in a latent-pixel domain for its efficiency and low-level prior~\cite{LatentDiffusionModels,SD3}.
Unfortunately, these diffusion models can involve significant training costs.
While much work is focused on inference speed or distillation into smaller models, we consider those efforts orthogonal to our work and leave out the relevant literature for brevity's sake.

However, the text prompts that condition these models are finicky and inaccurate for conveying user intent~\cite{PromptEngineering}, especially in terms of style.
Although negative prompts provide additional control, they can interfere with the original prompt or even be ignored~\cite{NegativePrompts}, while still  being restricted in their expressiveness.
These difficulties imply a need for more expressive control, which we believe to be in images.
Therefore, we now discuss both image-based conditioning for diffusion models and diffusion models for image-to-image translation tasks.

\subsection{Image-based Control for Diffusion Models}

\textbf{InstantStyle}~\cite{InstantStyle} conditions the output on the style of an input image without training.
To do so, the CLIP space embedding of its textual description is subtracted from that of the image to obtain a ``style direction vector'' in CLIP space: the text prompt cross attention layers in some blocks are then extended to also attend to this style vector.
While this model is neither explicitly trained to model the conditioning process, its success is undeniable.
In our research, we have found that the InstantStyle method produces the best style conservation process using conditioned images.

\textbf{IPAdapter}~\cite{IPAdapter} comprises a small neural translator to project from the input image's embedding, such as from ViT-H/14~\cite{ViT-H/14CLIP} CLIP~\cite{OpenAI-CLIP}, onto the embedding space used by the text encoder; the network cross-attends to these novel embeddings in additional cross-attention layers similar to those of the text prompt, effectively enabling it to use an image as prompt input.
IPAdapter is trained to reproduce the input image exactly --- only ``by coincidence'' is the emergent behaviour of the model to flexibly transfer the style and content of the condition image to the output images.
Our system leverages the lessons learnt from IPAdapter and InstantStyle to make the conditioning explicit, but bottlenecked through the Encoder-Decoder.

\textbf{UnCLIP-based Approaches}~\cite{UnCLIP} retrain the base model to reproduce an image based on its CLIP embedding, similar to IPAdapter, but replace the text prompt with the image condition completely, attaining a single mode of control over the output.
Since the entire model is retrained for this purpose, it risks catastrophic forgetting of the original model's capabilities, and is incompatible with other residual changes to the base model such as LoRA's~\cite{LORA}: we instead prefer to use seperate control models in order to keep the base model intact.

\textbf{ControlNet}~\cite{ControlNet} functions by cloning the base model's encoder.
The clone's inputs are replaced with the conditioning image, and its outputs are added as residuals to the original model's hidden states~\cite{ControlNet,ControlNet-XS}.
ControlNet excels at pixel-aligned conditioning of the output; it is not very succesful at style alignment.
However, we have found that the residual way in which the ControlNet affects the base model's output \emph{is} amenable to style alignment: just through conditioning on a stylecode rather than an input image.

\newpage

\section{Method}\label{sec:method}%

\subsection{Preliminaries}\label{sec:preliminaries}%

Diffusion models~\cite{DiffusionTheory,DDPM,DDIM} iteratively reverse a diffusion process that gradually transforms an image into noise (typically white Gaussian noise).
We write the forward noise process as starting from the data distribution $\vec{z}_0 \sim p(\vec{z})$ and ending with pure noise samples $\vec{z}_T \sim \mathcal{N}(0, \mathbf{I})$, over the course of $T$ time steps.
The immediate forward process is formally specified as
\begin{equation}
    \vec{z}_{t+1} \sim p(\vec{z}_{t+1} \vert \vec{z}_{t}) = \mathcal{N}\left(\sqrt{\alpha_{t+1}}\vec{z}_{t}, (1-\alpha_{t+1}) \mathbf{I}\right),
\end{equation}
where $\alpha_t$ denotes the so-called noise schedule.
Given this forward process, the diffusion model is trained to model the immediate denoising distributions, noted as $\hat{p}(\vec{z}_t \vert \vec{z}_{t+1})$.
During training, time steps are randomly sampled, and we directly supervise $\hat{p}(\vec{z}_t \vert \vec{z}_{t+1})$ by first sampling $p(\vec{z}_t \vert \vec{z}_0), \vec{z}_0 \sim p(\vec{z})$ followed by $\vec{z}_{t+1} \sim p(\vec{z}_{t+1}\vert \vec{z}_t)$.
Luckily, the exponential nature of additive white Gaussian noise implies a closed-form direct conditional $\vec{z}_t \sim p(\vec{z}_{t} \vert \vec{z}_0)$ which means that sampling $p(\vec{z}_t \vert \vec{z}_0)$ is constant-cost in terms of $t$.
By iteratively running the diffusion model for subsequent time steps, we can sample from the full generative model, written as
\begin{equation}
    \hat{p}(\vec{z}_0 \vert \vec{z}_{T}) = \prod\limits_{T}^{1}\hat{p}(\vec{z}_t-1 \vert \vec{z}_{t}), \qquad\vec{z}_{T} \sim \mathcal{N}(0, \mathbf{I}).
\end{equation}

As completely unconditional sampling is not very useful, diffusion models are trained to condition the generative distribution on an auxiliary input text prompt $\mathcal{T}$, modeling instead $p(\vec{z}_0 \vert \vec{z}_T, \mathcal{T})$ (known as \emph{Classifier-Free Guidance}~\cite{CFG}).
In our case, we wish further controllability by additionally conditioning the generative model on a style $\mathcal{S}$ to model $p(\vec{z}_0 \vert \vec{z}_T, \mathcal{T}, \mathcal{S})$.
We wish to leverage a pre-trained text-conditioned model by reusing its weights and adding small modules.
The base model is kept frozen to preserve its generative performance and expressivity.

\subsubsection{IPAdapter}%

In IPAdapter, for example, cross-attention layer that attend to (a CLIP~\cite{OpenAI-CLIP} embedding of) the style image are added after every prompt cross-attention layer.
The model is then trained by first sampling a data element $\vec{z}_0$ and then setting $\mathcal{S}=\mathrm{CLIP}(\vec{z}_0)$.
\Ie{} the model is only supervised to reproduce the condition image exactly --- even though this is not the (only) intended use-case.
Although the model is never trained to produce images with a different caption than the condition image, it shows emergent capabilities to do exactly that: it tends to generate images with the same style, composition, and identities as the condition.
However, it lacks any controllability of these aspects, and sometimes fails any or all of these aspects, depending on the text prompt.

\newpage

\subsection{StyleCode / Image  Conditioning}%

As discussed in \cref{sec:related} and \cref{sec:preliminaries}, existing techniques that condition on images do not provide a clear way to easily share styles with others.
To enable this, we encode the style-defining image as a 20 digit base64 stylecode, which can then be used to condition the image generation.

\subsubsection{Model architecture}\label{sec:architecture}

Our model architecture is a combination of a basic attention based autoencoder pair and a controlnet-style UNet decoder for residually controlling the frozen base model.
As the style codes are in embedding space, a decoder-based Stylecode-conditioned model  suffices.




\begin{figure}
    \centering
    \includegraphics[width=0.9\linewidth]{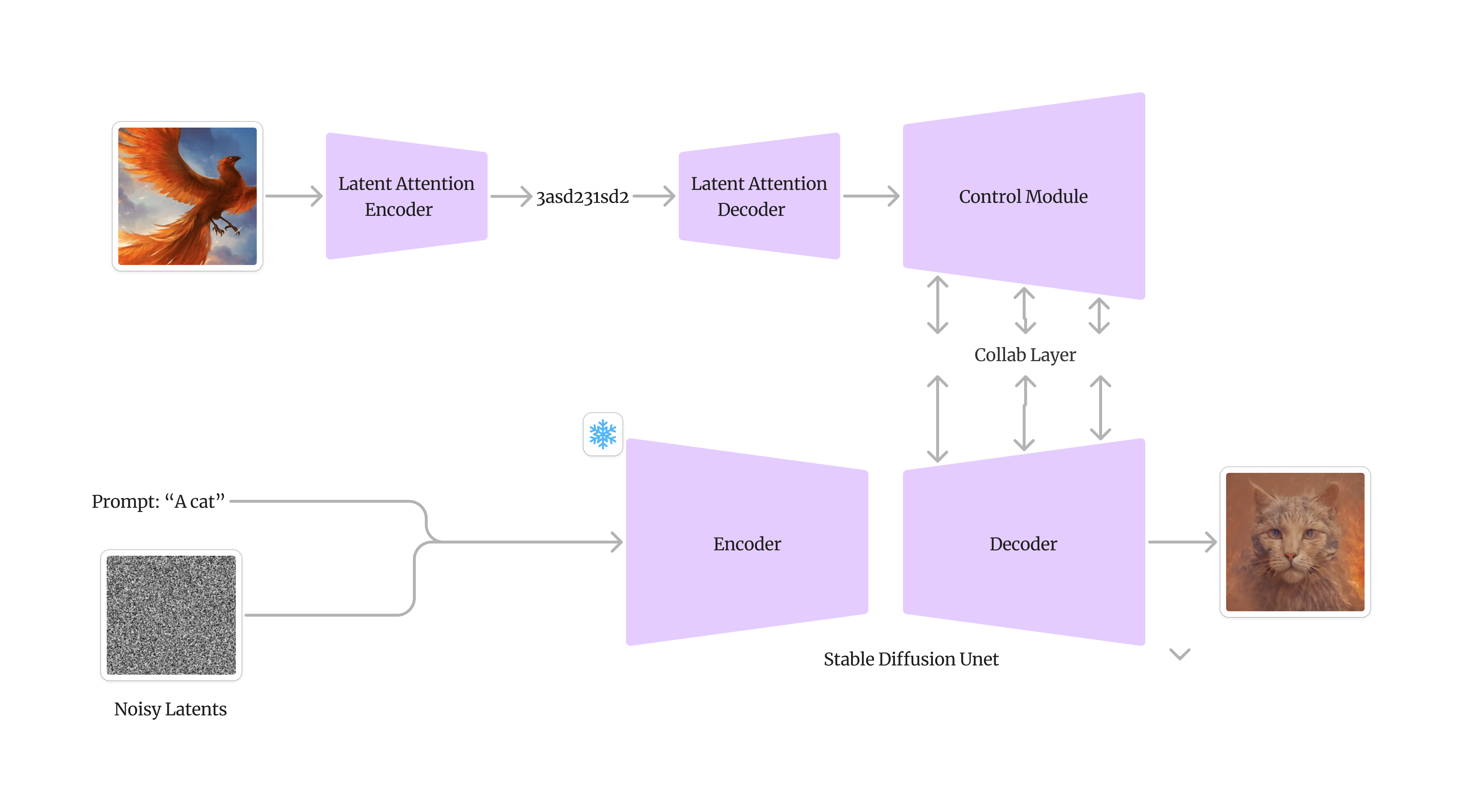}
    \caption{Auto Encoder and Control Module Architecture}
    \label{fig:enter-label}
\end{figure}

\subsubsection{AutoEncoder}
For our system, we used an latent autoencoder, 
The Encoder works by using 3 attention layers to attend from a high channel representation of the latent  to the embeddings, the latent is then projected down to the 20-dimensional latent size for the stylecode.

Later ,the Decoder takes this latent, makes a new output latent of the same shape as the midblock states for the Styleref Control Module and attends to the passed-in stylecode latent through its 3 layers.

During training to avoid discretization issues, the base64 encoding/decoding step is skipped, so the basic output latent is simply passed to the decoder. This is done because it allows you to use the model to train the latent used for the final stylecode, while ensuring a consistent flow of gradients during backpropagation.
This is trained jointly with the Stylecode-conditioned model to ensure the closest match between desired results and outputs.

\subsubsection{Stylecode}
For embedding the 20 dimensional latent, we chose to use base64 encoding due to it's widespread adoption and reliability, reducing any potential issues with character decoding later.
To do so, we quantize each of the 20 dimensions independently.
We also added a single number to the end of the stylecode, meant to denote the current stylecode encoder version. This way you can know if the stylecode you are looking at was made for a different decoder at a glance.
 
\subsubsection{Stylecode-conditioned model }
For the Stylecode-conditioned model  interaction with the main UNet, we used a decoder only version of the ControlNet scheme, residually affecting the internal hidden states of the base model.
In order to maximize the performance of the Stylecode-conditioned model , it consumes the internal state of the base model at every communication point (much like Collaborative Control~cite{CollaborativeControl}, except that our secondary model does not produce any output).

\subsubsection{Image Encoder}
For our Image encoder, we chose to use SigLip [need citation] due to it's superior performance over CLIP. It provides effective and efficient image embeddings and in our experiments trained drastically faster than simply passing the image wholesale into the Style Encoder .

\subsection{Dataset Generation}\label{sec:datasets}

For our dataset generation, we relied on InstantStyle~\cite{InstantStyle} , combined with source images from the MidJourney ~\cite{Midjourney} dataset and \cite{CommonCanvas }CommonCanvas, which were used as conditions for InstantStyle~\cite{InstantStyle} running with  SDXL~\cite{SDXL} IP Adapter \cite{IPAdapter}   and with prompts from a random image in the JourneyDB dataset \cite{JourneyDB}. This resulted in 35,000 condition, style and prompt dataset entries at a resolution of 1024x1024, which we downscaled to 512x512 for training (the native resolution of the base model).

\subsection{Training Process}

Our training process follows a simple training procedure with the base model frozen and the new layers initialized to white noise ($\sigma=10^{-4}$).
The residual layers are zero-initialized to ensure the base model is unaffected on initialization.
The base model of choice is StableDiffusion 1.5~\cite{LatentDiffusionModels} for its excellent balance of output diversity, controllability and accessibility --- it remains a staple model in the community for these reasons.
We use a batch size of $25$ and a learning rate of $10^{-6}$ for a  total of $100000$ steps, unless specified differently in the experiments.

\newpage
\section{Results}\label{sec:results}

 As shown in \cref{fig:enter-label}, style is effectively enforced from the stylecode with this method.
\begin{figure}
    \centering
    \includegraphics[width=0.75\linewidth]{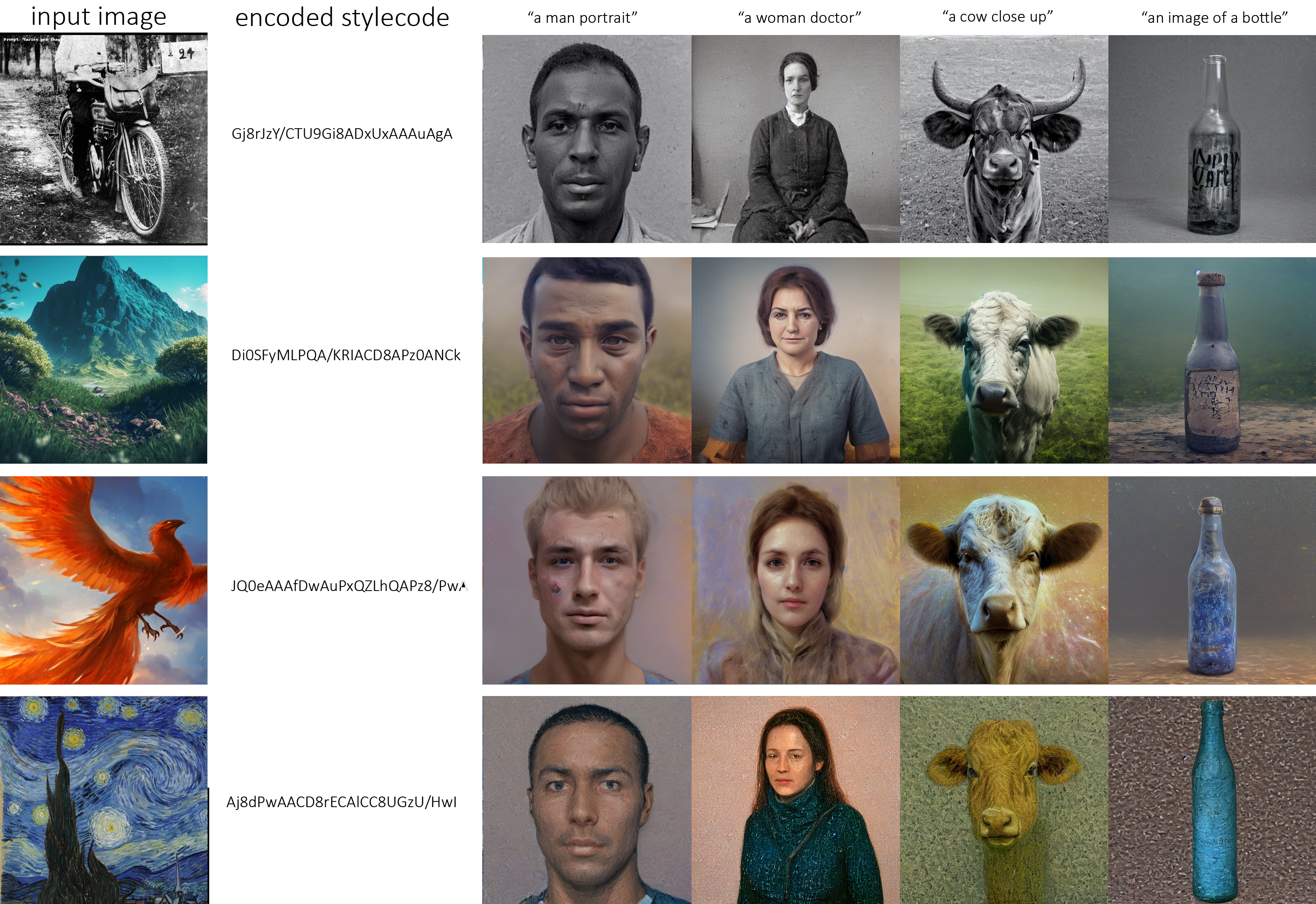}
    \caption{Example results with the left-most column being the source image with prompts "a close up man", "a woman portrait", "a cow" and "a bottle on a desk" with the same seeds after passing through the encoder to a stylecode and then used to generate the images.}
    \label{fig:enter-label}
\end{figure}

\begin{figure}
    \centering
    \includegraphics[width=0.75\linewidth]{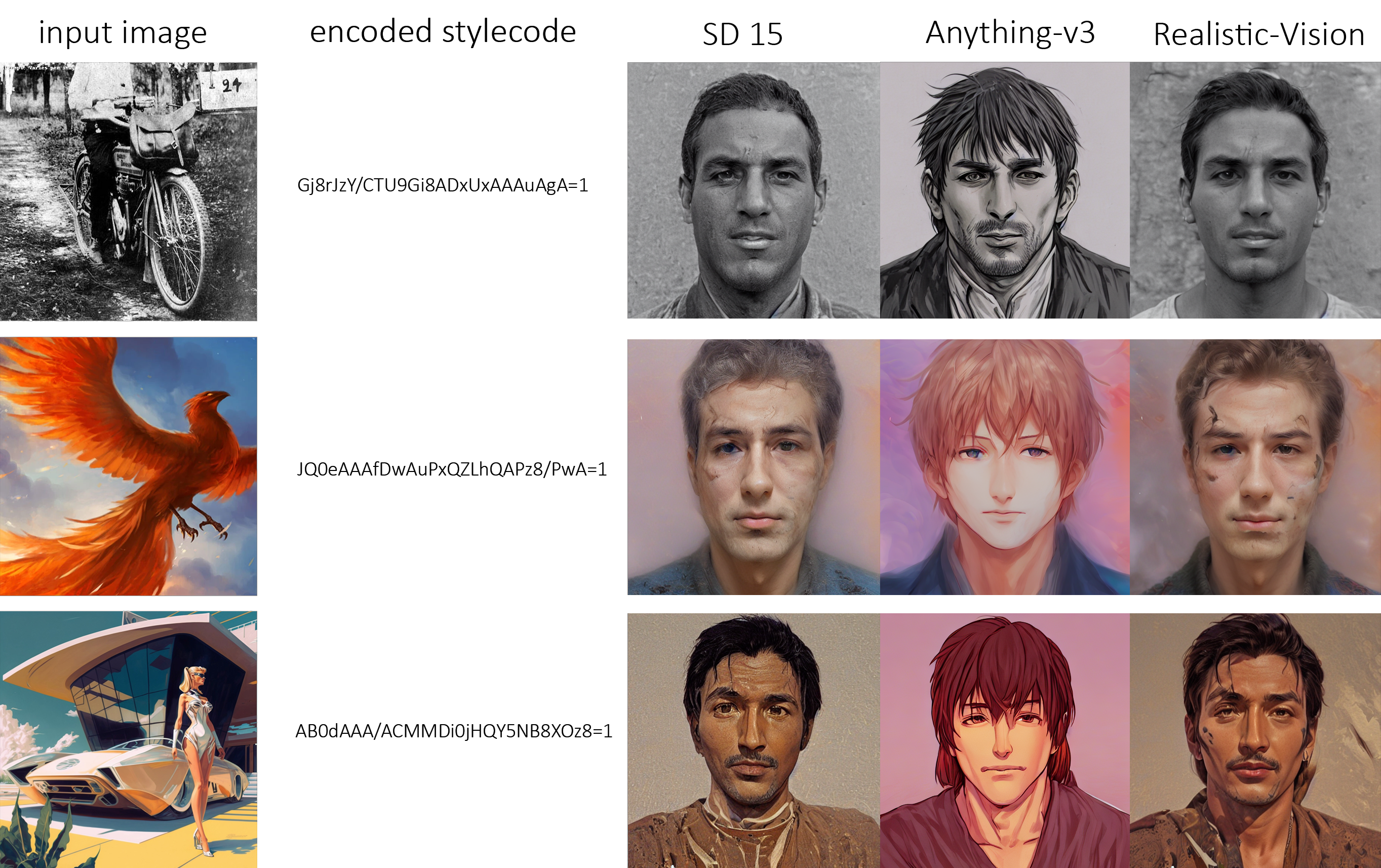}
    \caption{Example of the results with various trained base models and the control module with the prompt "portrait of a man" and four different style images}
    \label{fig:enter-label2}
\end{figure}

An additional benefit shown in \cref{fig:enter-label2} of the architecture is that due to the frozen base model, it can be switched out with fine-tuned variants with minimal performance degradation.


\newpage
\section{Conclusion, Limitations, and Future Work}\label{sec:conclusion}%

In this work, we have introduced stylecodes in order to allow social methods of control for image generation diffusion models conditioned on style images. By introducing this simple code condition, it is possible to easily share style information with friends about work done and use this to control local image generation models. 
This is all while fully preserving the functionality of the base model and it's compatibility with other control solutions like Loras and IP-Adapter.
We note that this partially replicates, but significantly expands, the functionality of MidJourney's srefs.
Furthermore, we openly publish our findings as well as source code, both for training and for inference.

We found that the main limitation was the training cost for the control model.
While it is relatively low for Stable Diffusion 1.5, larger DiT based models quickly become unwieldy and costly.
We also found that the dataset biased the distribution of the output model significantly, resulting in too narrow of a range of results, as well as issues with generations being reinforced in the final model.
Using a combination of synthetic and real data may be a better solution to this in future models.

We feel that this model and architecture will lead to more practical and sociable methods of control, allowing collaborative image generation between multiple people.
In terms of future work, we highlight the interplay with CFG, and specifically the option for multiple guidance~\cite{brooks2023instructpix2pix}. Additionally, using larger models and a more varied dataset will almost certainly improve the diversity of styles that can be created.

\bibliographystyle{splncs04}
\bibliography{main,add}

\end{document}